\documentclass[10pt,conference]{IEEEtran}

\IEEEoverridecommandlockouts
\usepackage[utf8]{inputenc}
\usepackage{cite}
\usepackage[inline]{enumitem}
\usepackage{booktabs}
\usepackage{graphicx}
\usepackage{balance}
\usepackage{epsfig}
\usepackage{epstopdf}
\usepackage{textgreek}
\usepackage{enumitem}
\usepackage{wrapfig}
\usepackage{amsmath}
\usepackage{amssymb}
\usepackage{amsthm}
\usepackage{algorithm}
\usepackage{algpseudocode}

\usepackage{bm}
\usepackage{url}
\usepackage{epstopdf}

\usepackage{amsfonts}

\ifCLASSINFOpdf
\else 
\fi
\usepackage{xcolor}
\usepackage[most]{tcolorbox}
\usepackage[x11names]{xcolor} 
\usepackage{amsmath}
\usepackage{enumitem}
\definecolor{darkgreen}{RGB}{0,100,0}

\usepackage{listings}
\definecolor{mymauve}{rgb}{0.58, 0.44, 0.86}
\lstdefinelanguage{json}{
  basicstyle=\ttfamily\footnotesize,
  breaklines=true,
  morekeywords={},
  showstringspaces=false,
  stringstyle=\color{blue},
  keywordstyle=\color{mymauve}\bfseries,
  commentstyle=\color{green},
  morestring=[b]"
}

\usepackage{lipsum}
\ifCLASSOPTIONcompsoc
    \usepackage[caption=false, font=normalsize, labelfont=sf, textfont=sf]{subfig}
\else
\usepackage[caption=false, font=footnotesize]{subfig}
\fi

\usepackage{times}
\usepackage[T1]{fontenc}
\usepackage{inconsolata}  
\usepackage{xcolor}
\usepackage{listings}
\usepackage{xcolor}
\usepackage{fancyvrb}
\definecolor{DodgerBlue}{RGB}{30,144,255}
\definecolor{ForestGreen}{RGB}{34,139,34}
\definecolor{Teal}{RGB}{0,128,128}
\definecolor{Orange}{RGB}{255,165,0}

\usepackage{soul}
\usepackage{multirow}
\usepackage{booktabs}
\usepackage{tabularx}
\usepackage{tikz}
\usepackage{tikzpagenodes}
\usetikzlibrary{calc}

\AddToHook{shipout/foreground}{%
	\ifnum\value{page}=1
	\begin{tikzpicture}[remember picture, overlay]
		\node[
		draw,
		text width=0.95\textwidth,
		align=left,
		font=\footnotesize,
		inner sep=5pt
		]
		at ($(current page header area.center) + (0,-5pt)$)
		{%
			This paper has been accepted for publication in the proceedings
			of the IEEE International Conference on
			Mobile Ad-Hoc and Smart Systems
			(MASS), 2026. This is the authors' accepted version
			of the article. The final version published by IEEE is:
		  A.~Alchaab, A.~Younis and D.~Pompili, ``X-RIC: An Agentic RAN-Intelligent Controller
		  Benchmark for Enhancing Reinforcement Learning,'' in \textit{Proc. of
				the IEEE International Conference on
				Mobile Ad-Hoc and Smart Systems
				(MASS)}, Hong Kong SAR, China, October 21--23, 2026.
		};
	\end{tikzpicture}%
	\fi
}

\begin{document}
\setlength{\dbltextfloatsep}{8pt plus 2pt minus 4pt}
\setlength{\textfloatsep}{8pt plus 2pt minus 4pt}
\title{{X-RIC: An Agentic RAN-Intelligent Controller Benchmark for Enhancing Reinforcement Learning}}

\author{\textbf{Adhwaa Al-Chaab†, Ayman Younis, Dario Pompili†}\\{†Department of Electrical and Computer Engineering}\\
{Rutgers University--New Brunswick, NJ, USA}\\
{E-mails: \{adhwaa.alchaab, a.younis, pompili\}@rutgers.edu
}\vspace{-.1in}}
\maketitle
\thispagestyle{empty}
\pagestyle{empty}



\begin{abstract}
\noindent The O-RAN ALLIANCE introduces architectural flexibility through disaggregation and AI-driven control. While the RAN Intelligent Controller (RIC) enables centralized optimization, current AI approaches lack reasoning capabilities and explainability. We propose eXplainable RIC (X-RIC), a novel framework integrating agentic AI into O-RAN architectures. X-RIC combines reactive Reinforcement Learning (RL) with strategic agentic reasoning through: (i) automated Behavioral Cloning (BC) using domain heuristics for sample-efficient training, (ii) a Telecom Language Model (TLM), a transformer-based reasoning module providing contextual abstraction and interpretability, and (iii) closed-loop co-adaptation within the Near-RT RIC control loop enabling continuous policy refinement. To ensure deployment readiness, we introduce BARX (Benchmarking Agentic Reasoning for RAN), a domain-specific evaluation framework assessing agentic control across latency, compliance, and determinism metrics. We validate X-RIC on SLA-aware network slicing in simulation and O-RAN-compliant testbed deployments. X-RIC achieves 89\% operational cost reduction while maintaining SLA compliance, converging 75\% faster with 50\% fewer training steps and 46\% lower variance.
\end{abstract}

\section{Introduction} 
The O-RAN ALLIANCE enables flexible, multi-vendor RAN ecosystems through open interfaces and programmable control~\cite{oran2023vertical}.  Central to this architecture, the RAN Intelligent Controller (RIC) provides near-real-time closed-loop optimization for diverse service types, including eMBB, URLLC, mMTC, and ISAC~\cite{lu2024integrated}. The O-RAN paradigm integrates AI directly into the control loop, requiring autonomous adaptation, real-time responsiveness, and intelligent decision-making under uncertainty~\cite{10944629}. However, realizing this AI-native vision faces several fundamental challenges~\cite{10872859,alchaab2025slice}, including: (i)~\emph{real-time responsiveness}, requiring decisions within strict latency bounds (sub-1s for Near-RT RIC) while adapting to dynamic conditions; (ii)~\emph{self-optimization under uncertainty}, as static policies cannot scale across heterogeneous network conditions; and (iii)~\emph{explainability and trust}, since AI-driven decisions must exhibit transparent behavior to satisfy operational and regulatory mandates. Traditional rule-based and supervised learning methods struggle with dynamic networking environments, while Reinforcement Learning~(RL), despite offering promise through sequential decision-making, suffers from fundamental limitations: (i)~\emph{poor sample efficiency}, requiring millions of interactions infeasible in live networks; (ii)~\emph{limited generalization} under unseen traffic patterns; and (iii)~\emph{lack of interpretability}, problematic for safety-critical operations~\cite{10114063}.

To address these limitations, this paper introduces \textbf{X-RIC}, a novel RIC framework integrating explainable agentic intelligence for AI-native RAN control. Unlike conventional RL approaches relying solely on reactive policies, X-RIC employs a hybrid architecture combining fast reactive learning with strategic reasoning. Built on agentic AI principles---augmenting learning systems with reasoning, temporal memory, and task-level understanding~\cite{10962241}---X-RIC enables contextual abstraction, reflective learning, and decision traceability.

The X-RIC architecture comprises three components: (i)~\emph{automated demonstration harvesting} that initializes RL policies through Behavior Cloning~(BC)~\cite{torabi2018behavioral} using trajectories from a fine-tuned reasoning model, improving sample efficiency without manual supervision; (ii)~a \emph{transformer-based reasoning module} providing strategic oversight and natural language explanations; and (iii)~\emph{closed-loop co-adaptation} enabling joint optimization while maintaining transparency. To further address sample inefficiency, X-RIC employs a \emph{dual-filtered harvesting pipeline}: episode-level policy-impact selection identifies valuable trajectories, while critic-guided pruning filters suboptimal actions.

\textbf{Related Work.} 
We review here the O-RAN architecture and how RL has been used within O-RAN, followed by online learning and LLM-based approaches.

\underline{\emph{O-RAN Architectures:}}
Recent O-RAN developments demonstrate AI integration potential~\cite{nvidia2025airan}. EdgeRIC~\cite{ko2024edgeric} enables TTI-level decision-making through near-RT applications interfacing with RAN infrastructure.  To enhance granular control,~\cite{10273394} embeds Central Units (CUs) and Distributed Units (DUs) within the real-time RIC as dedicated zApps for MAC scheduling and beam shaping. TinyRIC~\cite{liu2023tinyric} operates near CU/DU components, hosting tApps for non-3GPP interface management and energy optimization. While advancing O-RAN programmability, these frameworks lack integrated AI reasoning and autonomous adaptation. 
\indent\underline{\emph{RL in O-RAN:}}
RL has been widely used for intelligent resource management~\cite{10723502}, dynamic network slicing~\cite{10329913}, and closed-loop RIC control~\cite{9814869}. Younis et al.~\cite{10723502} propose a multitask DDPG framework for joint user association and power allocation, while~\cite{10329913} develops a multi-agent RL strategy for radio resource allocation across heterogeneous slices. Polese et al.~\cite{9814869} present a pipeline for training and validating deep RL-driven closed-loop control in O-RAN deployments. Despite strong performance, these RL methods remain constrained under highly dynamic conditions due to limited live-network exploration and scalability challenges in concurrent multi-slice deployments~\cite{10004901}.

\underline{\emph{Online Learning and LLM-Based Approaches:}}
To narrow the simulation-to-reality gap in network control, recent work has explored online learning and Large Language Model (LLM) integration. Liu et al. propose EdgeSlice~\cite{9355609}, a decentralized deep RL framework for jointly orchestrating radio and computing resources under slice-level SLAs. For network automation, ALLSTAR~\cite{elkael2025allstar} generates and validates RAN schedulers from high-level operator intents using LLMs. Other LLM-based efforts include TSpec~\cite{nikbakht2024tspec}, TelecomGPT~\cite{zou2024telecomgpt}, slice-aware resource allocation~\cite{wu2025llm}, and intent-based deployment~\cite{tang2025end}. However, existing online-learning methods still rely on fixed neural architectures, while current LLM-based solutions often operate as static inference engines without continuous adaptation, closed-loop validation, or latency guarantees suitable for sub-1s RIC control.

\underline{\emph{BC Approaches:}}
Behavioral Cloning (BC) has emerged as a promising alternative to address RL's sample efficiency by learning from expert demonstrations~\cite{torabi2018behavioral}, with applications in automated driving~\cite{9761743} and UAV detection~\cite{10975037}. However, existing BC approaches require extensive manual demonstrations that are hard to collect at scale and lack reasoning integration.

\textbf{Our Approach.} 
We address these limitations through three key aspects: (i)~automated demonstration collection using dual-filtering BC to generate high-quality training data without manual intervention, (ii)~integration of BC with transformer-based reasoning modules that provide natural-language rationales for explainable control, and (iii)~joint optimization of BC policies with agentic reasoning components for continuous adaptation in dynamic O-RAN environments. \emph{To the best of our knowledge, our work is the first to integrate automated BC within the O-RAN architecture for agentic RAN control.}

\begin{figure}[t]
\centering
\includegraphics[width=0.48\textwidth]{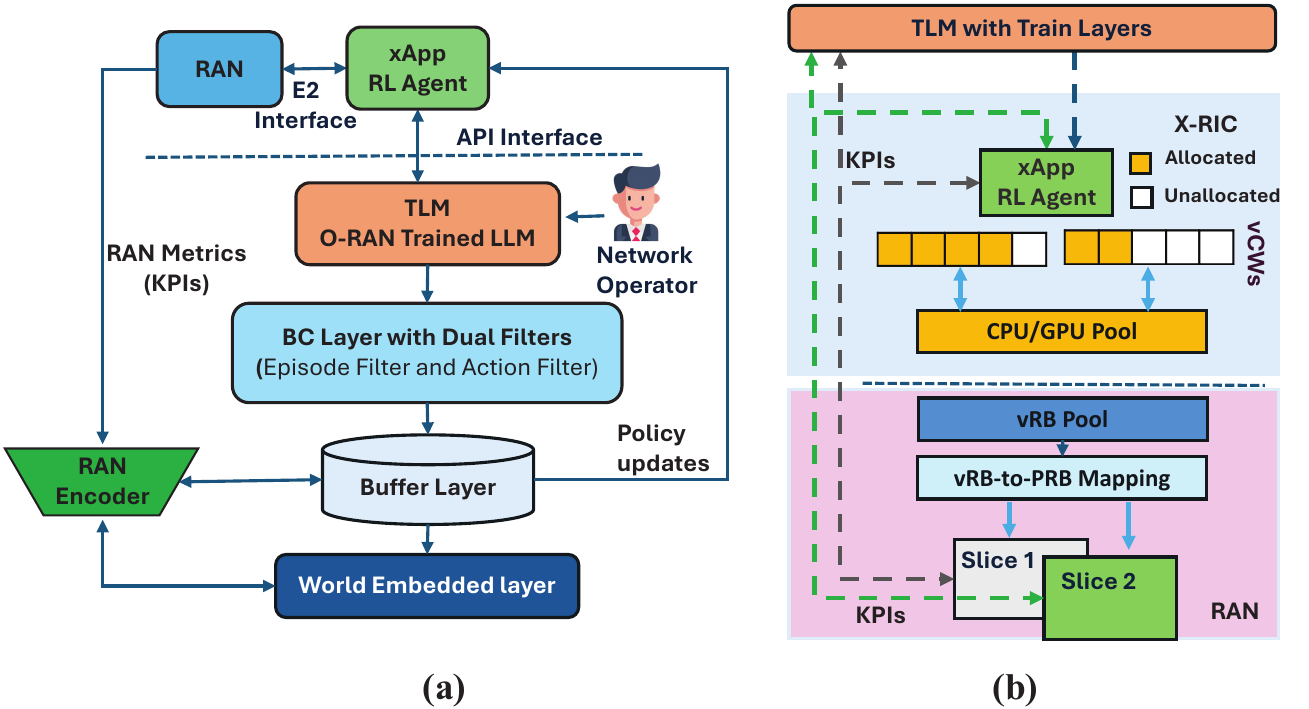}
\caption{X-RIC Framework architecture and O-RAN resource allocation: (a) Integrated TLM-RL system with dual-filtering BC, and (b) Virtual resource management for network slicing.}\label{fig:SM}
\end{figure}

\textbf{Our Contributions.}
The contributions of this paper are as follows: \textbf{\underline{i)}}~We design a novel eXplainable RAN intelligent controller, X-RIC, that integrates agentic AI with RL in a hybrid control architecture to enable sub-second decision-making based on natural-language intents. \textbf{\underline{ii)}}~We introduce a domain-specific evaluation framework, BARX, for assessing the suitability of agentic reasoning models in closed-loop telecom control.
\textbf{\underline{iii)}}~We propose a dual-filtering BC mechanism with automated demonstration harvesting, using episode-level policy-impact gating and action-level critic validation to improve sample efficiency while ensuring deployment-grade reliability. \textbf{\underline{iv)}}~As a case study, we formulate a network slicing resource allocation problem minimizing operational cost while meeting SLA requirements, 
achieving 89\% cost reduction versus baselines under Near-RT RIC 
timing constraints.

\textbf{Paper Organization.}
Sect.~\ref{sect:FR} details the main framework design for proposed X-RIC framework. Sect.~\ref{sect:SM} presents a case study demonstrating how X-RIC addresses network slicing in O-RAN. Sect.~\ref{sect:EV} reports the performance evaluation results. Finally, Sect.~\ref{sect:Con} concludes the paper.

\section{X-RIC Framework Design}\label{sect:FR}
We present the X-RIC framework integrating agentic reasoning into Near-RT RIC operations, introduce BARX for agentic RAN benchmarking, and describe dual-filtering demonstration collection 
for sample-efficient BC training.

\subsection{TLM Module and Benchmarking}

Our X-RIC framework, shown in Fig.~\ref{fig:SM}(a), is integrated within the O-RAN near-RT RIC; see~\cite{10024837} for an architectural overview. To avoid the computational overhead of full LLM fine-tuning, X-RIC uses a Telecom Language Model (TLM) architecture that augments a frozen pre-trained LLM with: (i)~domain-specific prompts encoding O-RAN operational knowledge, and (ii)~a trainable RAN-state encoder that maps structured telemetry, including KPIs, slice states, and resource metrics, into 256-dimensional embeddings. The encoder is trained via contrastive alignment on state-text pairs to project RAN observations into the LLM representation space, enabling RAN-aware reasoning without full model fine-tuning.

Integrating agentic reasoning modules into O-RAN workflows introduces system-level challenges due to the mismatch between probabilistic language model outputs and deterministic telecom control requirements. Four critical challenges emerge: (i)~\emph{Missed scheduling deadlines} when TLM inference time exceeds the $10~\rm{ms}$ to $1\rm{s}$ latency budget for Near-RT RIC operations~\cite{10024837}, (ii)~\emph{Malformed control messages} due to syntactically invalid JSON/YAML configurations, (iii)~\emph{Non-deterministic behavior} when identical network states produce different TLM outputs due to sampling randomness, and (iv)~\emph{Unintended SLA degradation} when semantically plausible but operationally suboptimal TLM decisions reduce network performance.
To address these challenges, we introduce a novel domain-specific framework, \emph{Benchmarking Agentic Reasoning~(BARX)}, based on O-RAN specifications. Unlike standard NLP benchmarks~\cite{srivastava2023beyond}, which emphasize linguistic fluency and semantic understanding, BARX evaluates agentic reasoning modules in closed-loop control settings. It assesses TLM candidates across four dimensions: latency, compliance, determinism, and policy impact.

For a given network state $s$, the TLM is queried $N$ times under fixed sampling parameters, temperature $T \in [0,1]$ and nucleus probability $p \in [0,1]$, yielding structured responses $\mathcal{E} = \{e_n\}_{n=1}^{N}$. \emph{i) Latency} is measured by the end-to-end response time $\ell_n$, from which we report the mean $\mu_\ell$, standard deviation $\sigma_\ell$, and empirical 95th percentile $\ell_{95}$. Responses with $\ell_n > 600\,\text{ms}$ violate the sub-second budget of Near-RT RIC control loops. \emph{ii) Compliance} is enforced by validating each reply against a strict Draft-7 JSON schema. The schema error rate $E$ is the fraction of invalid outputs; if $E>0$, BARX assigns a suitability score of 0. \emph{iii) Determinism} is quantified by pairwise comparison of valid responses, with consistency score $C$ defined as the proportion of identical reply pairs. Scores $C<1$ indicate non-deterministic behavior due to sampling noise or soft hallucinations. \emph{iv) Policy impact} is assessed by parsing valid replies into control directives executed under a TLM-guided Near-RT RIC policy.
The average episodic reward under TLM guidance, $\bar{R}_{\text{TLM}}$, is compared with the RL-only baseline reward $\bar{R}_{\text{RL}}$. The difference $\Delta \bar{R}$ quantifies the functional impact of agentic reasoning in the control loop. BARX aggregates $(\mu_\ell, \sigma_\ell, \ell_{95}, E, C, \Delta \bar{R})$ into a normalized score $[0,100]$, with thresholds calibrated using foundation-model evaluation practices~\cite{liang2022holistic}. Scores above 80 indicate deployment-grade performance; scores from 60 to 79 indicate marginal reliability requiring monitoring; scores from 40 to 59 require refinement; and scores below 40 are unsafe for production. Algorithm~\ref{alg:barx} describes the BARX procedure.

\subsection{X-RIC Agent System Architecture}\label{sub:xric_arch}

Our X-RIC framework in Fig.~\ref{fig:SM}(a) employs two coordinated agents with critic-guided selection for autonomous slice management. 
\emph{\underline{\textbf{Coordination Architecture.}}} The architecture consists of: (1)~the \emph{Monitoring Agent}, implemented as the TLM (based on LLaMA3-8B with LangChain tools), which continuously observes network performance using a two-tier temporal strategy: 1~s polling frequency for real-time monitoring from the Kinetica time-series database and 10~s aggregation windows for statistical smoothing. This design implements the periodic intervention strategy validated as optimal in Section~\ref{sect:ablation}.  The TLM monitors network state continuously and generates supervisory resource allocation recommendations $a_{\text{TLM}}$ every 10~s.(2)~The \emph{Control Agent}, implemented as a BC-trained actor-critic policy $\pi_{\theta_\pi}$ with critic network $Q_{\theta_Q}$, serves as the primary low-latency controller ($<10$~ms) for resource allocation. \emph{\underline{\textbf{Action Selection and Demonstration Harvesting.}}} At each 10~s decision epoch, both agents observe the same network state $s_k$: the Control Agent proposes $a_{\text{RL}}=\pi_\theta(s_k)$ while the TLM proposes $a_{\text{TLM}}$. The critic evaluates both candidates and executes the TLM recommendation when $\Delta R_k = Q_\phi(s_k,a_{\text{TLM}})-Q_\phi(s_k,a_{\text{RL}}) > 0$; otherwise, the RL action proceeds autonomously as the primary real-time controller. A dual-filter agentic BC layer, detailed in Sect.~\ref{sub:C}, generates high-quality demonstrations from positive TLM interactions, storing only critic-validated state-action pairs in the demonstration buffer $\mathcal{D}_{\text{fil}}$ for behavioral cloning updates. Policy updates flow back to the RL agent with minimal overhead (median TLM latency: 465.6~ms vs.\ sub-10~ms RL latency), forming a closed-loop learning system.


\subsection{Automated Demonstration Collection for BC}\label{sub:C}
Demonstration-augmented RL has shown promise in improving sample efficiency, particularly in complex decision-making tasks~\cite{ma2022comprehensive}. However, it still faces key limitations: expert demonstrations are often collected through manual teleoperation, or scripted policies. Although helpful in learning complex behaviors, the acquisition of high-quality demonstrations is often costly and impractical to scale~\cite{baker2022video, li2021meta}. To address these limitations, we propose \emph{automated demonstration harvesting using a TLM with episode-level policy-impact gating $\Delta\bar{R}_k$ and critic-validated action pruning}. Our approach integrates BC as imitation learning to enable agents to mimic high-quality TLM-guided behaviors, accelerating learning by bootstrapping policies with expert-like actions. Episodes with positive $\Delta\bar{R}_k$ values indicate improved performance from the use of agentic reasoning in the control loop, qualifying them as candidates for demonstration collection. However, these demonstrations may still contain suboptimal actions due to partial observability or ambiguous instruction mappings.
\begin{algorithm}[h!]
\footnotesize
\caption{BARX procedure}
\label{alg:barx}
\begin{algorithmic}[1]
\Require TLM model $M$, state $s$, parameters $(T,p)$,
queries $N$, JSON schema $S$
\Ensure score $\in[0,100]$, metrics
$(\mu_{\ell},\sigma_{\ell},\ell_{95},E,C,\Delta\bar{R})$
\State $\mathcal{E},\mathcal{L},V\gets\emptyset$
\Comment{Responses, latencies, valid responses}
\For{$n\gets1$ \textbf{to} $N$}
  \State $t_{\mathrm{start}}\gets\mathrm{getCurrentTime}()$;\ 
         $e_{n}\gets M.\mathrm{generate}(s,T,p)$;\ 
         $\ell_{n}\gets\mathrm{getCurrentTime}()-t_{\mathrm{start}}$
  \State $\mathcal{E}\gets\mathcal{E}\cup\{e_{n}\}$;\ 
         $\mathcal{L}\gets\mathcal{L}\cup\{\ell_{n}\}$
  \If{$\mathrm{validateDraft7Schema}(e_{n},S)$}\ 
         $V\gets V\cup\{e_{n}\}$
  \EndIf
\EndFor
\State $(\mu_{\ell},\sigma_{\ell},\ell_{95})\gets
       \mathrm{computeStats}(\mathcal{L})$
\State $E\gets1-\tfrac{|V|}{N}$
\Comment{Schema error rate}
\State $C\gets\tfrac{\mathrm{computePairwiseIdentical}(V)}
       {\mathrm{totalPairs}(V)}$
\Comment{Consistency score}
\State $\bar{R}_{\mathrm{TLM}}\gets\mathrm{executePolicy}(V)$;\ 
       $\Delta\bar{R}\gets\bar{R}_{\mathrm{TLM}}-\bar{R}_{\mathrm{RL}}$
\Comment{Policy impact}
\If{$E>0$}\ \Return $0,(\mu_{\ell},\sigma_{\ell},\ell_{95},E,C,\Delta\bar{R})$
\Else
  \State $score\gets\mathrm{normalizeMetrics}(
         \mu_{\ell},\sigma_{\ell},\ell_{95},E,C,\Delta\bar{R})$
  \State \Return $score,(\mu_{\ell},\sigma_{\ell},\ell_{95},
         E,C,\Delta\bar{R})$
\EndIf
\end{algorithmic}
\end{algorithm}

\begin{figure}[t]
\centering
\includegraphics[width=0.99\columnwidth]{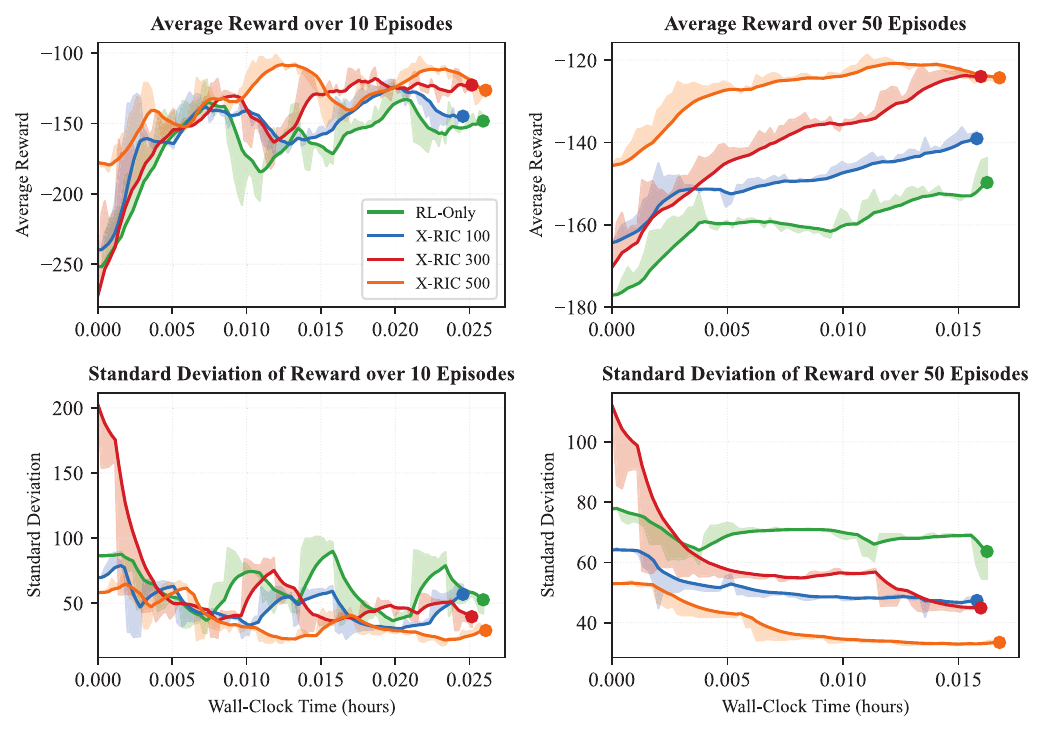}
\caption{Wall-clock learning curves comparing RL vs. X-RIC 
with 100, 300, 500 demonstrations (3 seeds, 10/50 episodes).}
\label{fig:a2c}
\vspace{-0.15in}
\end{figure}

\begin{figure}[t]
\centering
\includegraphics[width=\columnwidth, height=6.6cm]{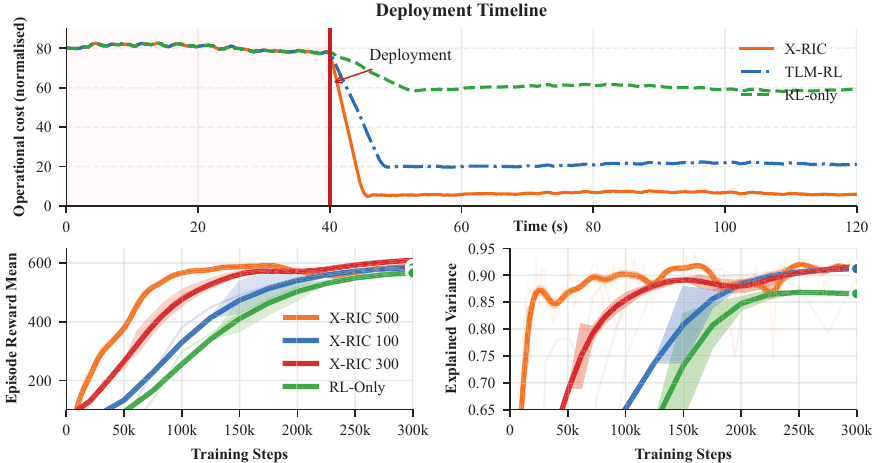}
\caption{Deployment cost (top), episode reward (bottom left), 
and explained variance (bottom right) comparing X-RIC variants 
with RL and TLM-RL baselines.}
\label{fig:a3c}
\end{figure}

\textbf{Dual-Filtering Mechanism.} 
Blindly cloning all actions from episodes with positive $\Delta \bar{R}_k$ values could reinforce suboptimal behaviors and lead to performance degradation. To address this, we propose a \emph{dual-filtering mechanism} that automates demonstration 
collection and retains only the most informative state--action pairs for training. The first filter operates at the \emph{episode level using a policy-impact metric, while the second operates at the action level using a critic-guided Q-filter.} This design addresses two critical challenges in demonstration-augmented RL; (i) reducing the cost of demonstration collection by leveraging TLM-generated trajectories, and (ii) improving demonstration quality through critic-guided filtering of suboptimal actions.

\paragraph{Episode-level filtering} Per-step reward differences from the outset would require comparing all steps across both TLM and RL runs—potentially thousands of comparisons—many of which may derive from episodes offering no benefit. To measure the impact of the TLM-guided controller during early training, we calculate $\Delta \bar{R}_k = \bar{R}_{\text{TLM},k} - \bar{R}_{\text{RL},k}$ for episode $k$, where $\bar{R}_{\text{TLM},k} = \frac{1}{T_k} \sum_{t=0}^{T_k-1} r_{k,t}^{\text{TLM}}$, with $r_{k,t}^{\text{TLM}}$ representing the reward at step $t$ in episode $k$ and $T_k$ denoting the number of steps in episode $k$; $\bar{R}_{\text{RL},k}$ is computed analogously. The metric $\Delta \bar{R}_k$ serves as an initial coarse filter to identify promising episodes. This episode-level comparison provides a fast and reliable method to assess whether the TLM enhances overall performance across full control sessions, effectively smoothing out noisy or unstable per-step rewards that may arise from random network events (e.g., traffic spikes or signal interference). By computing a single score per episode, we can discard unhelpful episodes, thereby avoiding the computational cost of analyzing every individual step. Episodes with $\Delta \bar{R}_k > \delta$ (margin $\delta \ge 0$) are admitted into a raw pool $D_{\text{raw}}$, assuming they encapsulate meaningful policy improvements. This step removes trajectories where agentic reasoning underperforms while maintaining full automation.


\paragraph{Action-level filtering}
Episode-level selection may include trajectories with mixed-quality decisions. We therefore retain a state-action pair $(s_k, a_k)$ only if the critic prefers the demonstrator's action to the current policy's proposal,
\begin{equation}
Q(s_k, a_k | \theta_Q) > Q(s_k, \pi(s_k | \theta_\pi) | \theta_Q) + \epsilon.
\end{equation}
with optional margin $\epsilon \geq 0$. Here, $a_k$ is the TLM-guided action, $\pi(s_k | \theta_\pi)$ is the current policy, and $Q(\cdot | \theta_Q)$ is the critic with parameters $\theta_Q$. This Q-filter removes suboptimal actions and yields a distilled demonstration buffer $\mathcal{D}_{\text{fil}}$ suitable for imitation.


\textbf{Q-filtered BC Objective.}
Given $\mathcal{D}_{\text{fil}}$, we define a selective BC loss,
\begin{equation}\label{eq:LBC}
\begin{split}
\mathcal{L}_{\text{BC}} &= \frac{1}{|\mathcal{D}_{\text{fil}}|}
\sum_{(s_k,a_k) \in \mathcal{D}_{\text{fil}}}
\|\pi(s_k | \theta_\pi) - a_k\|_2^2 \\
&\quad \cdot \mathbb{I}\left[
Q(s_k, a_k | \theta_Q) >
Q(s_k, \pi(s_k | \theta_\pi) | \theta_Q) + \epsilon
\right],
\end{split}
\end{equation}
where $|\mathcal{D}_{\text{fil}}|$ is the number of retained samples and $\mathbb{I}[\cdot]$ is the indicator function. The actor update direction balances RL and selective imitation,
\begin{equation}
g_{\theta_\pi} =
\lambda_1 \nabla_{\theta_\pi} J(\theta_\pi)
- \lambda_2 \nabla_{\theta_\pi} \mathcal{L}_{\text{BC}},
\quad \lambda_1, \lambda_2 > 0.
\end{equation}
where $J(\theta_\pi)$ is the standard return-maximization objective (e.g., DDPG policy gradient). We anneal $\lambda_2$ as the policy approaches demonstrator performance, preventing the actor from being constrained below expert level. 

In Fig.~\ref{fig:a2c}, performance trends are tracked in real time to assess system-level efficiency, showing that demonstration-augmented X-RIC improves sample efficiency: X-RIC$_{500}$ achieves a $3\times$ faster reward gain and 46\% lower variance than the RL baseline. In Fig.~\ref{fig:a3c}, training and deployment metrics are tracked to assess operational efficiency, showing that X-RIC$_{500}$ achieves $\sim$75\% faster convergence to comparable reward, 0.90 explained variance with 50\% fewer training steps (vs.\ RL-only at $\sim$0.87), and $\sim$6$\times$ lower operational deployment cost than the RL and TLM-RL baselines.

\section{SLA Guarantee in O-RAN Use Cases}\label{sect:SM}
In this section, we apply our proposed X-RIC framework to the network slicing setup~\cite{alchaab2025slice,11027044}, which involves dynamic configuration and poses significant challenges for traditional RL methods. Our objective is to minimize the total operating cost of maintaining multiple slices while ensuring that each slice meets its SLA-defined performance requirements.

\subsection{Problem Description}
As shown in Fig.~\ref{fig:SM}(b), we consider a single-cell O-RAN-compliant mobile network with a gNB supporting three slices, $i\in\mathcal{I}=\{\mathrm{eMBB}, \mathrm{URLLC}, \mathrm{mMTC}\}$. Each slice serves a subset of UEs, where $|\mathcal{U}^e|=N_e$, $|\mathcal{U}^u|=N_u$, and $|\mathcal{U}^m|=N_m$ denote the number of users in each slice. The network operates over discrete time slots $k \in \mathcal{K}$. An SLA is a contract between the Mobile Network Operator (MNO) and slice tenants specifying QoS requirements (e.g., maximum latency, minimum throughput). The MNO deploys an xApp within the Near-RT RIC to dynamically allocate two virtual-resource primitives~\cite{schmidt2021flexric}: i)~\emph{virtual Resource Blocks (vRBs)}, $b_{i,k}\in\mathbb{Z}_{\geq 0}$, representing discrete spectrum units assigned to slice~$i$; and ii)~\emph{virtual Compute Workloads (vCWs)}, $c_{i,k}\in [0,1]$, representing the CPU/GPU time-sharing fraction. Let $N_{\mathrm{vRB}}$ denote the total available vRBs and $C_{\mathrm{w}} = 1$ the normalized compute capacity. 
The achievable data rate for user $u$ in slice $i$ can be written as,
\begin{equation}\label{eq:rate_model}
\mu_{i,k,u} = \frac{b_{i,k}}{N_i} W_{\mathrm{vRB}} \log_2\left(1 + \gamma_{i,k,u}\right),
\end{equation}
where $W_{\mathrm{vRB}}$ is the bandwidth per vRB, $N_i = |\mathcal{U}^i|$, and $\gamma_{i,k,u}$ is the SNR. The latency for URLLC users is $L_{i,k,u} = L_{i,k,u}^{\mathrm{tx}} + \kappa/c_{i,k}$, where $\kappa > 0$ is a processing constant.

\textbf{KPI Definition.}
The slice-specific KPIs are defined as,
\begin{equation}\label{eq:kpi}
\small
\text{KPI}_{i,k} \!=\! 
\begin{cases} 
\frac{1}{N_e} \sum_{u \in \mathcal{U}^e} \mu_{i,k,u} & i = \text{eMBB}, \\[0.3em]
\frac{1}{N_m} \sum_{u \in \mathcal{U}^m} \mathbb{I}(\mu_{i,k,u} > \mu^{\mathrm{min}}_u) & i = \text{mMTC}, \\[0.3em]
\frac{1}{N_u} \sum_{u \in \mathcal{U}^u} \max\!\left(0, 1 - \frac{L_{i,k,u}}{L_{\max}}\right) \mu_{i,k,u} & i = \text{URLLC},
\end{cases} 
\end{equation}
where $\mathbb{I}(\cdot)$ is the indicator function, $\mu^{\mathrm{min}}_u$ is the minimum rate for mMTC, and $L_{\max}$ is the maximum allowable URLLC latency. The $\max(0,\cdot)$ operator ensures non-negative KPI contributions.

\textbf{Cost Model and Problem Formulation.}
The operating cost with unit prices $u_{\mathrm{vRB}}$ and $u_{\mathrm{w}}$ for vRBs and vCWs is formulated as,
\begin{equation}\label{eq:cost}
\small
 \text{Cost}(\mathbf{b}_k,\mathbf{c}_k)
    = u_{\mathrm{vRB}}\!\sum_{i\in\mathcal{I}} b_{i,k}
    + u_{\mathrm{w}}\!\sum_{i\in\mathcal{I}} c_{i,k}
    - \sum_{i\in\mathcal{I}} w_i\,\text{KPI}_{i,k},
\end{equation}
where $w_i > 0$ weights the KPI importance. The resource allocation problem is formulated as,
\begin{subequations}\label{eq:cost_dyn}
\begin{align}
\min_{\mathbf{b}_k, \mathbf{c}_k} \quad & \text{Cost}(\mathbf{b}_k, \mathbf{c}_k) \\
\text{s.t.}\quad 
& \text{KPI}_{i,k} \geq Q_i, \quad \forall i \in \mathcal{I}, \label{eq:sla}\\
& \sum_{i\in\mathcal{I}} b_{i,k} \leq N_{\mathrm{vRB}},\quad \sum_{i\in\mathcal{I}} c_{i,k} \leq 1, \label{eq:capacity}\\
& b_{i,k} \in \mathbb{Z}_{\geq 0}, \; c_{i,k} \in [0,1], \quad \forall i \in \mathcal{I}, \label{eq:domain}
\end{align}
\end{subequations}
where $Q_i$ is the SLA threshold for slice $i$. Problem~\eqref{eq:cost_dyn} is a mixed-integer nonlinear program (MINLP) due to integer vRBs and the nonlinear rate function~\eqref{eq:rate_model}. We adopt an RL approach to solve it in near real time.

\begin{figure}[t]
\centering
\includegraphics[width=0.5\columnwidth]{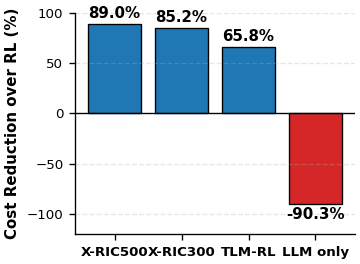}
\vspace{-0.1in}
\caption{Percentage cost reduction over RL for X-RIC and TLM-RL, compared with a non-fine-tuned LLM policy.}
\label{fig:cost_reduction}
\end{figure}

\subsection{MDP Model}
We formulate~\eqref{eq:cost_dyn} as a Markov Decision Process (MDP) with states $\mathcal{S}$, actions $\mathcal{A}$, and policy $\pi_\theta: \mathcal{S} \rightarrow \mathcal{A}$. The objective is to maximize $R = \sum_{k\in\mathcal{K}} \gamma^k r_k$, where $\gamma \in [0,1)$ is the discount factor.

\textbf{State Space.} The state at slot $k$ is,
\begin{equation}\label{eq:state}
s_k =\Bigl(\bigl\{\text{KPI}_{i,k-1}\bigr\}_{i\in\mathcal{I}}, \bigl\{\lambda_{i,k}\bigr\}_{i\in\mathcal{I}}, \bigl\{\rho_{i,k}\bigr\}_{i\in\mathcal{I}}\Bigr),
\end{equation}where $\lambda_{i,k}$ and $\rho_{i,k}$ denote the arrival rate and service demand characteristics for slice $i$, respectively.

\textbf{Action Space.}
The agent outputs continuous values $a_k = \{(\tilde{b}_{i,k}, c_{i,k})\}_{i\in\mathcal{I}}$, where $\tilde{b}_{i,k} \in [0, N_{\mathrm{vRB}}]$ and $c_{i,k} \in [0, 1]$. The discrete vRB allocation is obtained by rounding: $b_{i,k} = \lfloor \tilde{b}_{i,k} + 0.5 \rfloor$.

\textbf{Reward Function.} The reward incentivizes cost minimization with SLA compliance,
\begin{equation}\label{eq:reward}
R(s_k, a_k) = -\text{Cost}(\mathbf{b}_k, \mathbf{c}_k) - \eta \sum_{i \in \mathcal{I}} \max(0, Q_i - \text{KPI}_{i,k}),
\end{equation}where $\eta > 0$ penalizes SLA violations. 

Fig.~\ref{fig:cost_reduction} compares total cost reduction (vRB and vCW allocation) across four approaches, with RL as the reference point (0\%). X-RIC$_{500}$ achieves 89.0\% cost reduction, X-RIC$_{300}$ achieves 85.2\%, and TLM-RL achieves 65.8\%, showing that BC-trained policies outperform both baselines. The non-fine-tuned LLM increases cost by 90.3\% due to hallucinations that violate resource constraints, highlighting the need for domain-specific BC training.

\begin{figure}[t]
\centering
\includegraphics[
  width=0.48\textwidth,
  trim=0 0 0 1.55in,
  clip
]{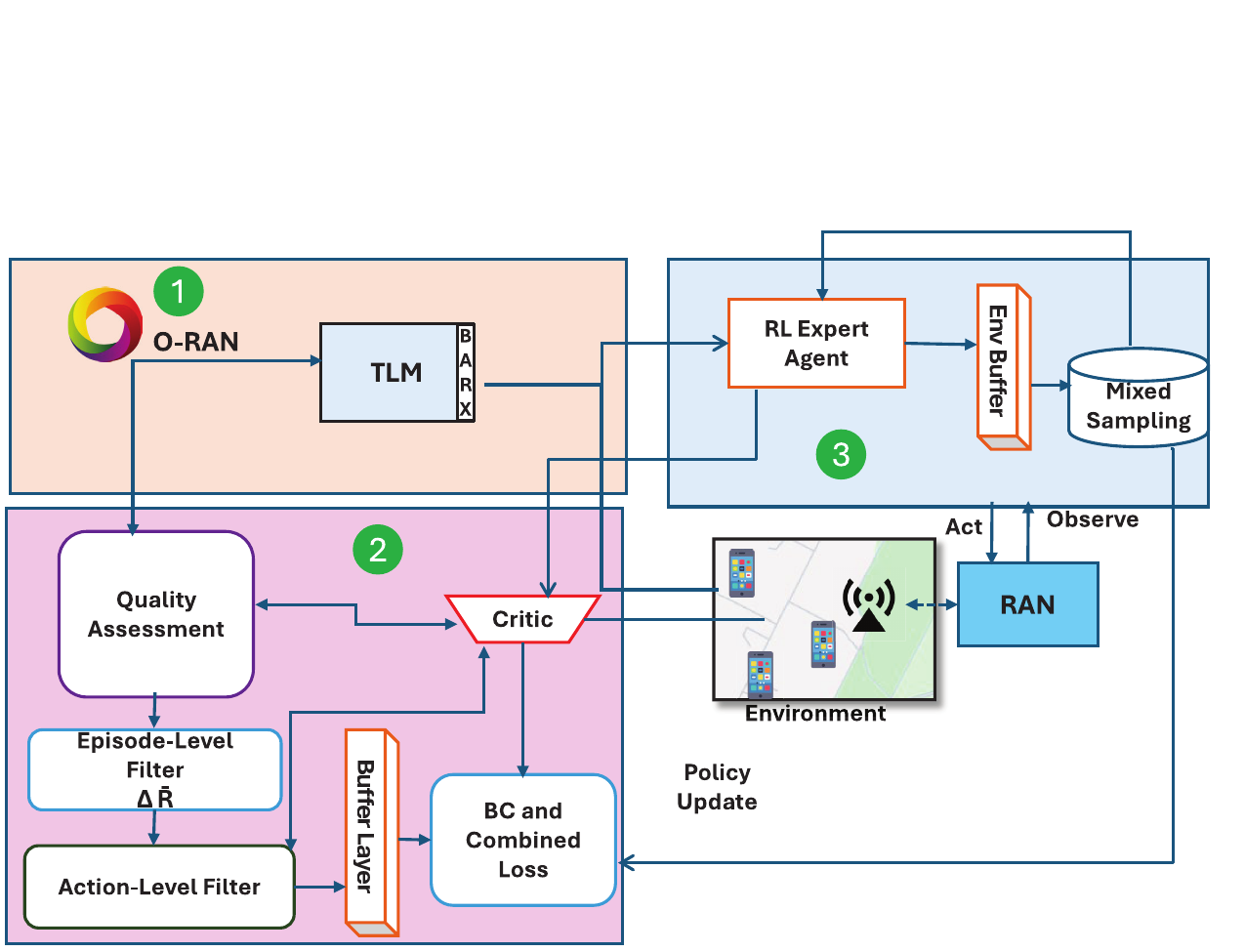}
\caption{TLM-guided RL agent with dual-filtering BC.}
\label{fig:step}
\end{figure}

\subsection{TLM-based RL Approach Solution}
In this subsection, we present a novel approach that integrates TLM feedback with RL for dynamic resource allocation in O-RAN network slicing, leveraging the dual-filtering BC mechanism detailed in Section~\ref{sub:C}. Our solution follows a systematic approach, as shown in Fig.~\ref{fig:step}, organized into the following steps,

\textbf{Step 1: Enhanced DDPG Architecture Design.}
We adopt the DDPG algorithm~\cite{lillicrap2015continuous} for its suitability in mixed-integer resource allocation problems where vRBs are discrete and vCWs are continuous. Additionally, DDPG’s off-policy nature enables learning from
data not necessarily generated by the current policy, such as
exploration trajectories or expert demonstrations. The architecture comprises an actor network $\pi_{\theta_\pi}$ and critic network $Q_{\theta_Q}$, with state observations $s_k$ encompassing slice KPIs, arrival rates, and service-demand characteristics as inputs.

\textbf{Step 2: Demonstration Collection and Filtering.}
We apply the dual-filtering mechanism described in Section~\ref{sub:C} to collect high-quality demonstrations from TLM interactions. This process generates the filtered demonstration buffer $\mathcal{D}_{\text{fil}}$ containing only high-quality state-action pairs that pass both episode-level policy impact gating and action-level critic validation. We enhance the standard DDPG objective by incorporating TLM guidance and BC losses,
\begin{equation}
\small
\mathcal{F}(\theta_\pi, \theta_Q) = -\mathcal{L}_p(\theta_\pi) + \lambda_v \mathcal{L}_v(\theta_Q) + \lambda_{\text{TLM}} \mathcal{L}_{\text{TLM}}(\theta_\pi) + \lambda_{\text{BC}} \mathcal{L}_{\text{BC}}(\theta_\pi)
\end{equation}
where the BC loss $\mathcal{L}_{\text{BC}}(\theta_\pi)$ utilizes the Q-filtered demonstrations from $\mathcal{D}_{\text{fil}}$ as defined in~\eqref{eq:LBC}.


\textbf{Step 3: Mixed Replay Training.}
During training, we sample from both the standard replay buffer $\mathcal{D}$ (exploration data) and filtered demonstration buffer $\mathcal{D}_{\text{fil}}$ (expert guidance). This mixed sampling strategy enables the agent to learn from both self-exploration and TLM-guided expert demonstrations.

\textbf{Step 4: State Encoding and TLM Integration.}
The RAN encoder transforms network states $s_k$ into textual representations
$\Omega_k$ for TLM-based reasoning over resource allocation. The TLM processes
structured prompts containing: (i)~the optimization objective, (ii)~the current
network observation, including per-slice PRB usage, buffered packets, and
traffic criticality indicators, and (iii)~an action-to-slice mapping over
candidate decisions. It then evaluates the RL agent's proposed action and
returns a structured JSON response with a reasonability judgment, explanation,
and corrected action when needed. Fig.~\ref{fig:llm_prompt_decision}
illustrates this process for an emergency URLLC traffic burst, where the TLM detects that allocating
PRBs to eMBB and mMTC is suboptimal under high URLLC congestion and overrides
it with a latency-aware allocation. The resulting decisions are filtered
through the dual-filtering BC mechanism and integrated into training.

\textbf{Step 5: Continuous Learning Loop.}
The system operates in a closed-loop manner where: (i)~X-RIC generates expert demonstrations based on current network conditions, (ii)~dual-filtering selects high-quality demonstrations, (iii)~BC loss guides policy learning from filtered demonstrations, (iv)~RL continues exploration and optimization, and (v)~performance feedback updates the X-RIC guidance process. This integrated approach utilizes the sample efficiency of BC while maintaining the exploration capabilities of reinforcement learning.

\begin{figure}[t]
\centering
\setlength{\fboxsep}{3pt}
\setlength{\fboxrule}{0.4pt}
\fbox{%
\begin{minipage}{\dimexpr0.88\columnwidth-2\fboxsep-2\fboxrule\relax}
{\fontsize{6.2}{7.0}\selectfont
\textbf{Prompt Instruction:} You are a network operator, and your goal is to devise a resource-
allocation policy that maximizes utility while ensuring that QoS targets for latency, throughput, and packet loss are met.

\medskip
\fbox{%
\begin{minipage}{\dimexpr\linewidth-2\fboxsep-2\fboxrule\relax}
{\fontsize{6.2}{7.0}\selectfont
\textbf{\textcolor{orange}{Observation:}} Step = 55,
Action = \texttt{[0]}\\
PRB Usage: \{URLLC: 10.5\%, eMBB: 9.3\%, mMTC: 3.5\%\}\\
Buffered Packets: \{URLLC: 16, eMBB: 7, mMTC: 3\}

\smallskip
\textbf{\textcolor{DodgerBlue}{Thoughts:}} Emergency
traffic burst detected in \textbf{URLLC}.

\smallskip
\textbf{Action-to-Slice Mapping:}
\begin{itemize}[leftmargin=*,nosep]
  \item \textbf{Action 0:} Allocate PRBs to both
        \textbf{eMBB} and \textbf{mMTC}.
  \item \textbf{Action 1:} Prioritize \textbf{URLLC}.
  \item \textbf{Action 2:} Distribute PRBs equally
        across all slices.
  \item \textbf{Action 3:} Prioritize \textbf{eMBB}.
\end{itemize}

Assess whether \textbf{Action 0} is appropriate given
congestion, PRB usage, and service priority.

\smallskip
\textbf{\textcolor{teal}{Final Decision:}}

\smallskip
\fbox{%
\begin{minipage}{\dimexpr\linewidth-2\fboxsep-2\fboxrule\relax}
{\fontsize{6.2}{7.0}\selectfont\ttfamily
\{"decision": "unreasonable",\\
\phantom{x}"explanations": "Action 0 is suboptimal.\\
\phantom{xx}URLLC shows high buffering and is\\
\phantom{xx}latency-critical, while eMBB and mMTC\\
\phantom{xx}have lower demand.",\\
\phantom{x}"final\_action": "1"\}
}
\end{minipage}%
}
}
\end{minipage}%
}
}
\end{minipage}%
}
\caption{LLM reasoning under emergency traffic.}
\label{fig:llm_prompt_decision}
\vspace{-0.12in}
\end{figure}
\begin{figure*}[!ht]
\centering
\begin{tabular}{ccc}
\hspace*{.3cm}\includegraphics[width=0.28\textwidth, height=1.6in]{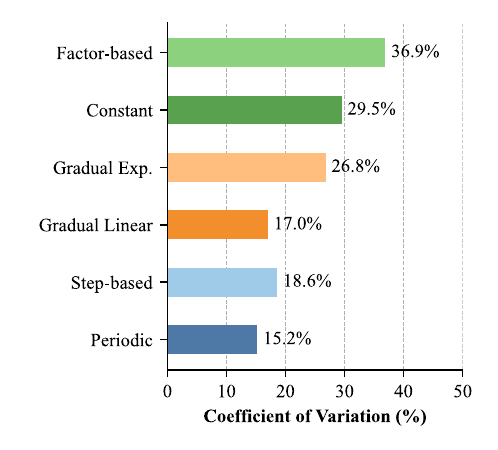} &
\hspace*{.3cm}\includegraphics[width=0.28\textwidth, height=1.6in]{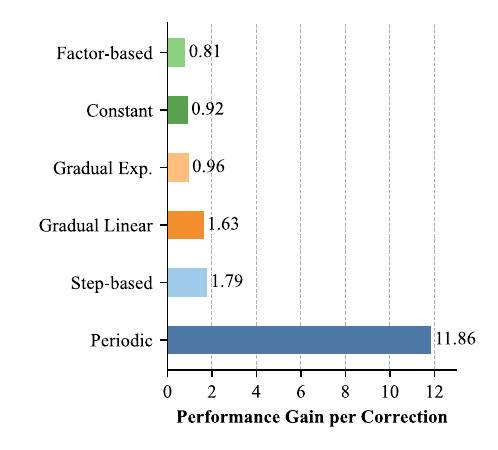} &
\hspace*{.3cm}\includegraphics[width=0.28\textwidth, height=1.6in]{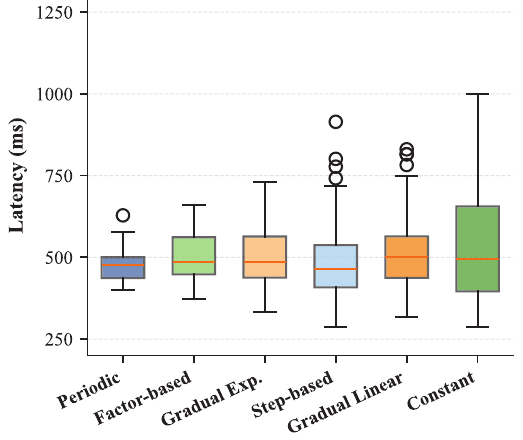} \\
\small (a) & \small(b) & \small(c)
\end{tabular}
\vspace{-0.1in}
\caption{a) Performance stability across intervention strategies, 
measured by coefficient of variation (lower is more stable), 
(b) Efficiency of correction, measured by performance gain per 
intervention, and (c) Latency distribution per strategy 
(5 episodes).}
\label{fig:ab2}
\vspace{-0.1in}
\end{figure*}
\begin{figure*}[!ht]
\centering
\begin{tabular}{ccc}
\hspace*{.3cm}\includegraphics[width=0.28\textwidth, height=1.4in]{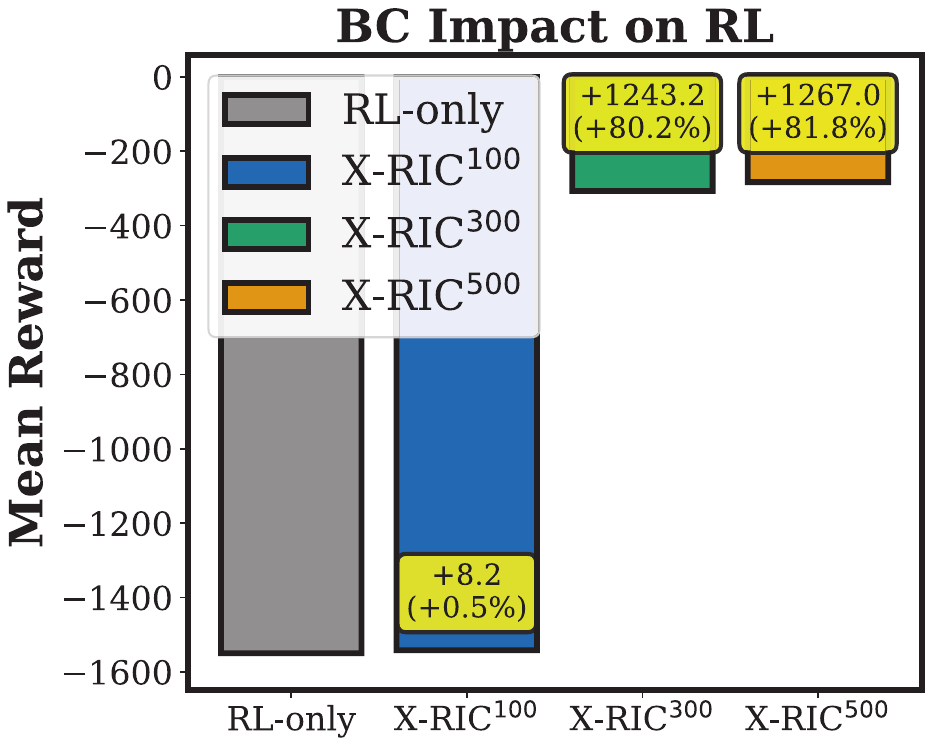} &
\hspace*{.3cm}\includegraphics[width=0.28\textwidth, height=1.4in]{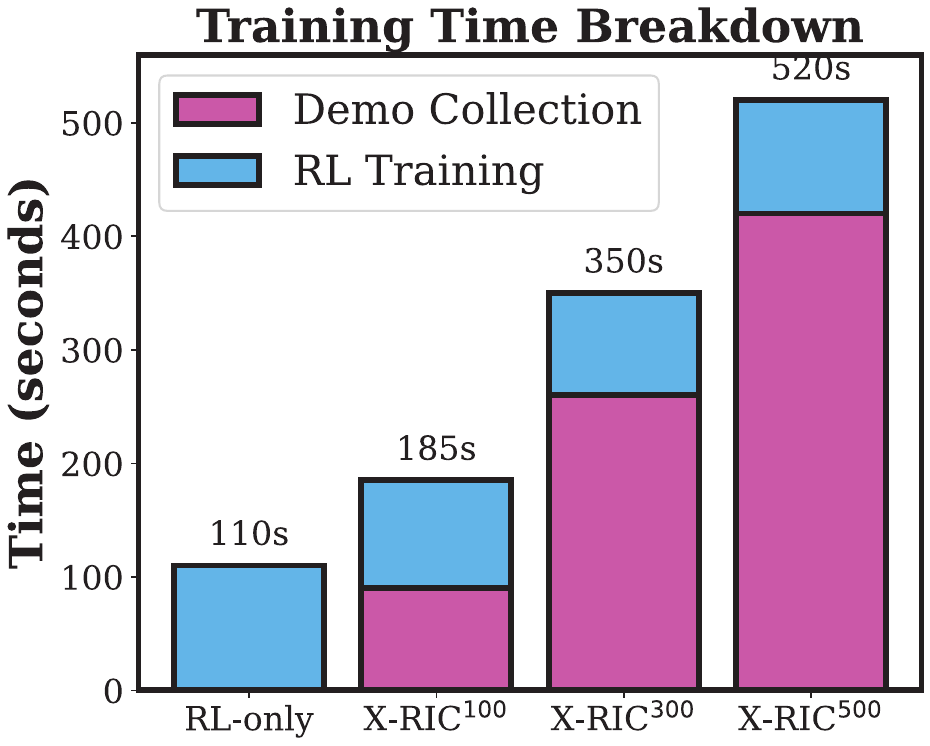} &
\hspace*{0.3cm}\includegraphics[width=0.28\textwidth, height=1.4in]{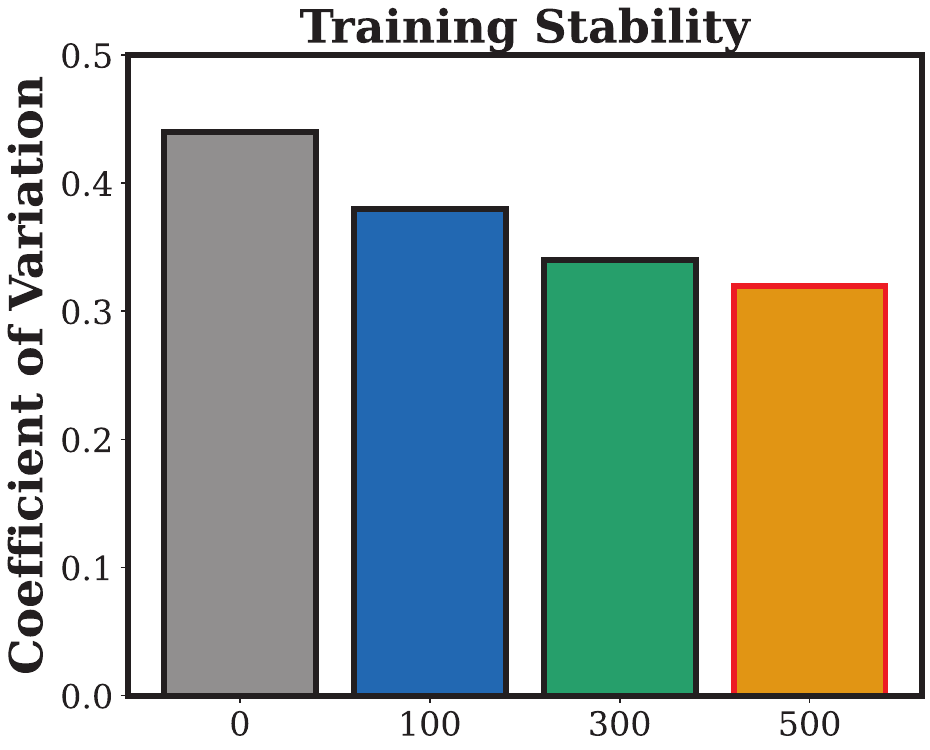} \\
\small (a) & \small(b) & \small(c)
\end{tabular}
\vspace{-0.1in}
\caption{Impact of BC on RL performance across X-RIC variants; 
(a)~Mean reward comparison, (b)~Training time breakdown, and 
(c)~Coefficient of variation indicating training stability.}
\label{fig:BC}
\vspace{-0.15in}
\end{figure*}

\begin{table}[t]
  \centering
  \scriptsize
  \renewcommand{\arraystretch}{0.95} 
  \setlength{\tabcolsep}{2pt} 
\caption{BARX results for LLM models.}
  \label{tab:barx_results}
  \begin{tabular}{@{}lrrrr@{}}
    \toprule
    Model & Latency (ms) & Compliance (\%) & Determinism (\%) & Policy Impact \\
    \midrule
    L3   & $441.9 \pm 45.9$   & 100.0   & 100 & $5.39$ \\
    Q7B  & $573.9 \pm 69.8$   & 100.0   & 100 & $6.09$ \\
    Q4B  & $459.7 \pm 67.0$   & 100.0   &  92 & $4.04$ \\
    QT   & $465.9 \pm 75.6$   & 83.3    &  83 & $3.09$ \\
    MT   & $588.9 \pm 152.6$  & 100.0   & 100 & $5.06$ \\
    TL   & $342.7 \pm 270.1$  & 0.0     &   0 & $-2.75$ \\
    DS   & $878.2 \pm 135.3$  & 100.0   & 100 & $5.88$ \\
    \bottomrule
  \end{tabular}
\end{table}
\section{Performance Evaluation}\label{sect:EV} 
All experiments are performed on a workstation equipped with an AMD Ryzen 9 9950X CPU (16C/32T, 5.75 GHz, AVX-512) and an RTX 5080 GPU (16 GB VRAM). The RL Control Agent is implemented as a BC-trained DDPG policy with vectorized state normalization, evaluated over 300-second episodes with sub-second control intervals under dynamic traffic patterns and congestion scenarios.

\subsection{TLM Module Benchmarking} Prior to deploying the agentic TLM model, we benchmark seven fine-tuned open-source LLMs via BARX to assess their suitability for agentic reasoning in O-RAN under realistic conditions: LLaMA3-8B (L3), Qwen-7B (Q7B), Qwen-4B (Q4B), Qwen:latest (QT), Mistral:latest (MT), TinyLLaMA:latest (TL), and DeepSeek-7B (DS). Each model is evaluated under identical traffic-control 
prompts with three sampling regimes: $(T,p) \in \{(0.01,0.8), (0.1,0.9), 
(0.0,1.0)\}$. For each model-regime combination, we conduct five independent trials with 20 queries each, yielding 100 JSON responses per model regime (2100 total responses). Table~\ref{tab:barx_results} summarizes the BARX results. Although TL exhibited the lowest latency ($342~\rm{ms}$), it produced $100\%$ parse errors from preambles and inconsistent key naming (e.g., \texttt{explanaition}), violating strict control-plane interfaces. Q4B occasionally omitted required fields (e.g., \texttt{``final\_action''}) without syntax errors, indicating latent schema risks and reducing determinism by $8\%$.  QT, with $T = 0.1$ and $p = 0.9$, produced consecutive JSONs, causing ``Extra data'' parse errors and resulting in a schema error rate of $E = 16.7\%$. The BARX score is calculated for each model by first enforcing two strict requirements: full schema compliance and a $\ell_{95}$ below $600\,\mathrm{ms}$. Models that fail either criterion receive a score of $0$. For models that qualify, The final score averages normalized metrics: $\mu_\ell$ (20\%), $\sigma_\ell$ (15\%), $\ell_{95}$ (15\%), $C$ (20\%), and $\Delta\bar{R}$ (30\%), eliminating 71\% of candidates.  L3 is selected as the base reasoning module in our experiments due to its favorable latency-reliability trade-off.


\subsection{Ablation of TLM Intervention Schedules}\label{sect:ablation}

To address the question \emph{when should the agentic reasoning agent intervene and correct the RL policy?}, we conducted an ablation study evaluating six distinct correction strategies. The experimental methodology compared baseline RL performance against six TLM schemes, each tuned for fairness and consistency. The strategies included: \emph{Constant Correction}, where the TLM continuously overrides RL decisions; \emph{Periodic Correction}, which applies guidance at fixed intervals (every $10~\rm{steps}$) with a sparse, consistent intervention rate (correction probability $\approx 0.1$); \emph{Step-Based Correction}, which intervenes only during the initial $1000~\rm{steps}$ before transitioning to full autonomy; \emph{Linear Decay}, which reduces the correction probability linearly from $1.0$ to $0.0$ over 3000 steps; \emph{Exponential Decay}, applying a decay rate of $0.95$ every 100 steps; and \emph{Factor-Based Correction}, which maintains a constant intervention probability of $\lambda = 0.5$. All strategies were evaluated over five episodes of $300$ simulation seconds using a pre-trained DDPG agent with vectorized normalization in a realistic O-RAN control environment. TLM-based corrections were applied using deterministic action predictions from our O-RAN–fine-tuned L3 model to ensure reproducibility. Our results, presented in Fig.~\ref{fig:ab2}, show that the periodic correction strategy achieved the best overall trade-off. It delivered the highest performance gain per correction, $11.86$, the lowest coefficient of variation, $15.2\%$, and a median latency of $465.6~\rm{ms}$. Based on these findings, we adopt periodic correction as the preferred TLM intervention method in X-RIC. 

\textbf{Simulation Setup.} For the SLA network slicing use case described in Section~\ref{sect:SM}, we generate three slices with $N_e = 10$, $N_u = 10$, and $N_m = 6$ users. The slice-specific SLA requirements follow Eq.~(\ref{eq:kpi}): $Q_{\mathrm{eMBB}}=Q_{\mathrm{URLLC}}=12$~Mbps, the mMTC KPI is the fraction of users satisfying $\mu_{i,k,u}>\mu_u^{\min}$, and $L_{\max}=10$~ms for URLLC. The total number of vRBs is $N_{\mathrm{vRB}} = 12$, with normalized compute capacity $C_w = 1$ and unit costs $u_{\mathrm{vRB}} = u_w = 1$ in~(\ref{eq:cost}). The SLA violation penalty is set to $\eta = 10$ in~(\ref{eq:reward}). The environment exposes standard RL interfaces (\texttt{reset}, \texttt{step}, and \texttt{observation}) with discrete and continuous control over user association. A TLM-assisted \texttt{VecEnv} wrapper supports explainability and human-in-the-loop policy refinement. We run 300-second episodes with Sub-second control intervals.
\begin{figure}[t]
\centering
\begin{minipage}{0.99\columnwidth}
\centering
\includegraphics[width=0.92\linewidth]{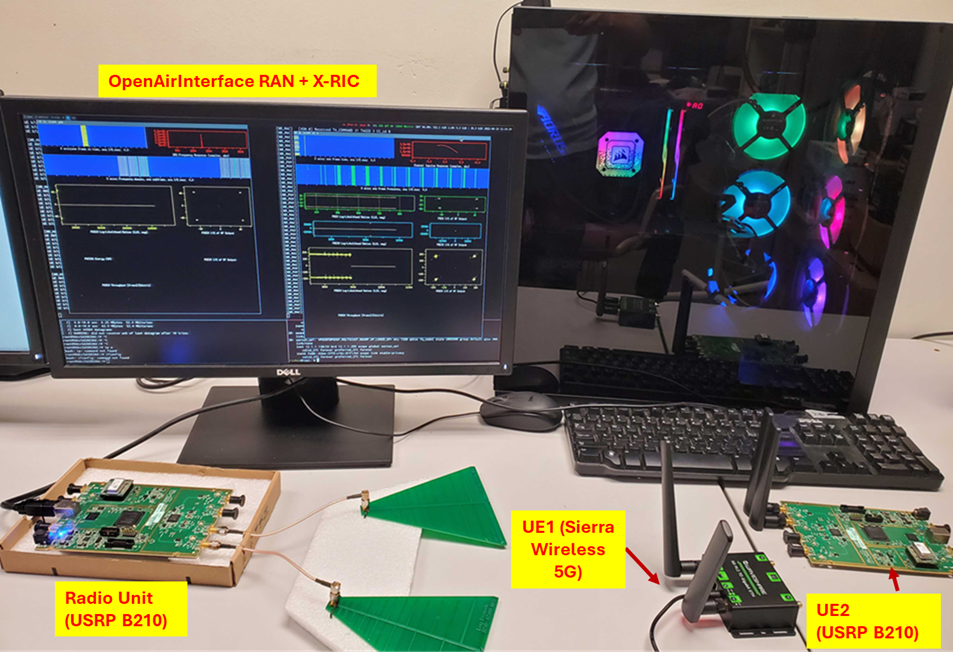}
\end{minipage}
\caption{O-RAN testbed deployment for X-RIC evaluation.}
\label{fig:testbed}
\end{figure}

\begin{figure*}[!ht]
\centering
\begin{tabular}{cccc}
\hspace*{-0.1cm}\includegraphics[width=0.23\textwidth, height=3.5cm]{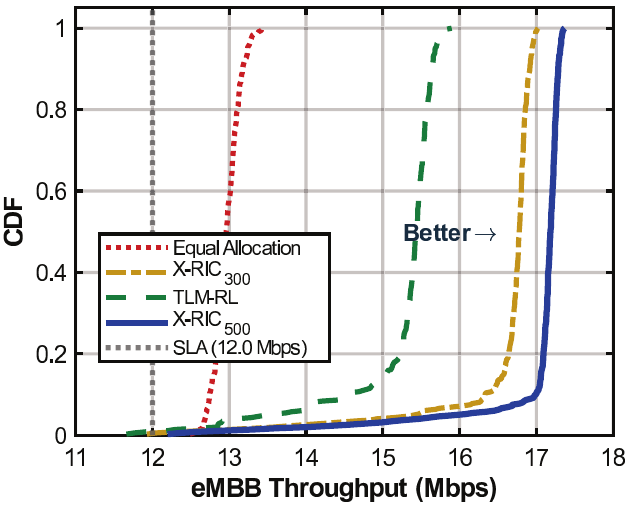} &
\hspace*{-0.05cm}\includegraphics[width=0.23\textwidth, height=3.5cm]{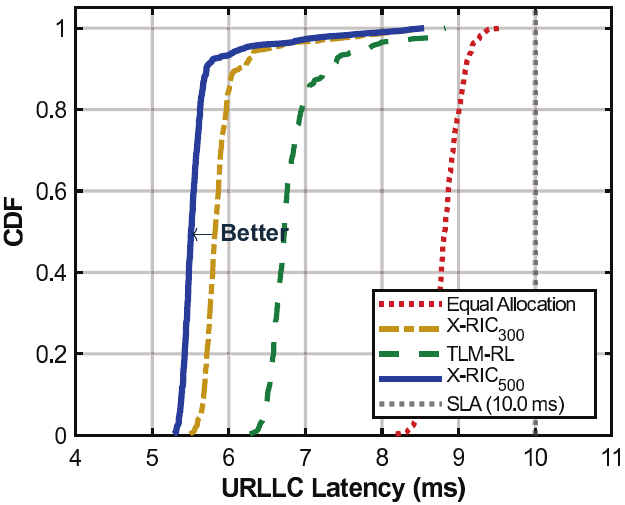} &
\hspace*{-0.1cm}\includegraphics[width=0.23\textwidth, height=3.5cm]{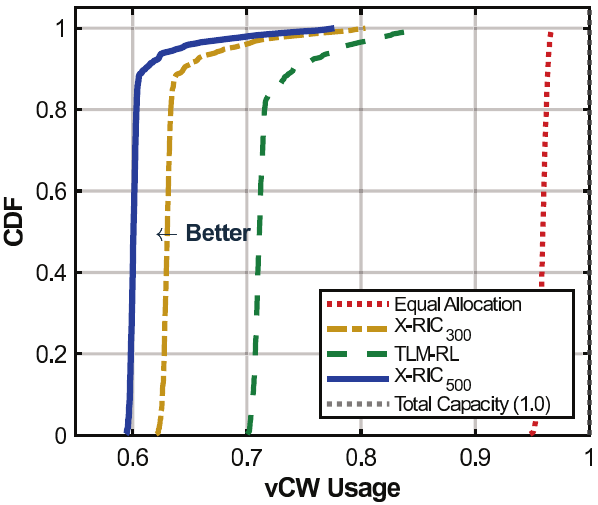} &
\hspace*{-0.1cm}\includegraphics[width=0.23\textwidth, height=3.5cm]{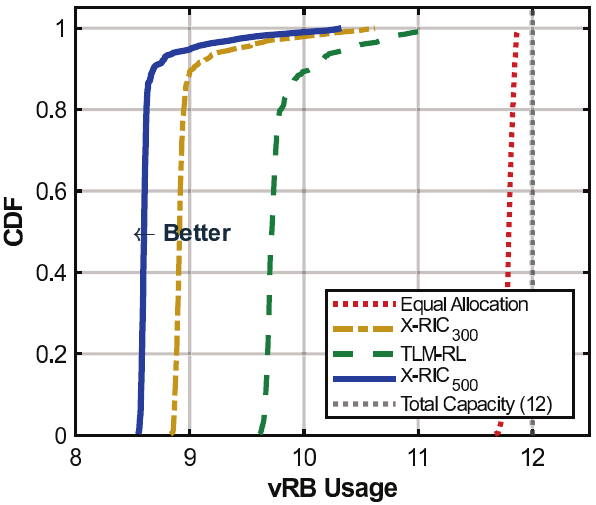}
\\
\small (a) & \small (b) & \small (c) & \small (d)
\end{tabular}
\vspace{-0.1in}
\caption{Over-the-air testbed evaluation of X-RIC for SLA-aware network slicing; (a) CDF of eMBB throughput across methods with 12.0 Mbps SLA threshold, (b) CDF of URLLC latency with 10.0 ms constraint, (c) vCW utilization efficiency relative to total compute capacity, and (d) vRB utilization efficiency relative to available spectrum resources.}
\label{fig:a3}
\end{figure*}

Fig.~\ref{fig:BC}(a) shows that 
increasing demonstration buffer size improves policy performance, with 
X-RIC~300 and X-RIC~500 achieving 80.2\% and 81.8\% reward improvements 
over RL baseline. These gains incur only moderate increases in training 
time (see Fig.~\ref{fig:BC}(b)), indicating high returns on demonstration 
effort. Fig.~\ref{fig:BC}(c) further reveals that higher-quality demonstrations 
consistently reduce the coefficient of variation, highlighting greater policy stability. RL often fails to achieve high returns and exhibits high 
variance~\cite{fujimoto2021minimalist}, which limits deployment reliability 
in real-time control systems. The dual-filtering mechanism addresses this 
limitation by selecting high-quality demonstrations that reduce learning instability and improve convergence. These results confirm that demonstration 
quality directly impacts both training efficiency and deployment stability.

\subsection{Over-the-Air Testbed Implementation}
To validate simulation results 
through OTA deployment, we implement a testbed with three main components 
(Fig.~\ref{fig:testbed}). \textbf{i)}~\emph{RAN and control:} OpenAirInterface 
gNB using USRP B210 SDR interfaces with FlexRIC near-RT RIC via E2 for 
closed-loop slice control. KPIs stream to a Kinetica real-time accelerated 
database; TLM is based on L3-8B with LangChain tools.\textbf{ii)}~\emph{5G core:} OAI 5G core (AMF, SMF, UPF) configured for network slicing and per-slice policy enforcement. \textbf{iii)}~\emph{5G UEs:} Sierra with programmable SIMs. 

\textbf{Over-the-Air Experimental Setup.}
We instantiate X-RIC with L3-8B as the TLM foundation and BC-trained DDPG 
as the RL Control Agent, implementing a two-slice subset: URLLC 
(Slice~1, UE1, 12.1.1.0/24) with strict QoS requirements (zero packet loss, 
low latency) and eMBB (Slice~2, UE2, 12.1.1.128/24) with high-throughput 
requirements. Following the objective in Section~\ref{sect:SM}, the operator intent is:
``Minimize total operating cost while maintaining SLA-defined QoS requirements: the URLLC slice must satisfy strict latency and reliability constraints; the eMBB slice must meet throughput targets, with resource allocation
prioritizing URLLC during congestion.'' X-RIC autonomously translates this intent into concrete controls: PRB allocations, 
compute workload distribution, and per-slice throughput guarantees. 

Initially, the network operates under balanced resource allocation with equal PRB distribution (50:50 ratio). To create realistic congestion scenarios, we use iperf3 in UDP mode to enable controlled congestion testing. Unlike TCP's automatic throttling, UDP transmits at configured bitrates regardless of capacity, allowing traffic demands to exceed available bandwidth and exposing packet loss for QoS evaluation. We deploy two iperf3 servers running on ports 5201 (Slice~1) and 5202 (Slice~2) to generate traffic demands exceeding the system capacity ($\sim$296~Mbps in our setup). The servers generate per-UE metrics (throughput, jitter, packet loss) at 1-second intervals, which are ingested into a Kinetica accelerated time-series database. When aggregate traffic demand exceeds available radio capacity, X-RIC detects congestion through KPI monitoring and evaluates SLA compliance across both slices. The critic-guided mechanism dynamically reallocates PRBs to restore SLA requirements with minimal resource consumption.

{\textbf{Performance Baselines.}} We evaluate the SLA compliance and resource efficiency of X-RIC against three representative baselines using Cumulative Distribution Function (CDF) analysis across four key metrics: eMBB throughput, URLLC latency, vCW utilization, and vRB utilization. We compare X-RIC against the following approaches: (i) \textit{Equal Allocation}, which distributes vRB and vCW resources uniformly across all slices without adaptive control, representing static provisioning strategies; (ii) \textit{TLM-RL}, which integrates TLM guidance with the RL agent but without the behavioral cloning mechanism, isolating the contribution of automated demonstration harvesting; and (iii) \textit{X-RIC$_{n}$}, our framework initialized with $n$ dual-filtered demonstrations to assess the impact of demonstration budget on performance. 

\emph{\underline{\textbf{SLA Compliance Analysis.}}} Fig.~\ref{fig:a3}(a)--(b) show eMBB throughput and URLLC latency vs.\ SLA requirements of 12.0~Mbps and 10.0~ms, respectively. Equal Allocation operates close to limits: eMBB at 12--13~Mbps and URLLC at 9--10~ms, risking violations during congestion. X-RIC$_{500}$ provides larger safety margins: eMBB at 17~Mbps (42\% above SLA) and URLLC at 5~ms (50\% below) with minimal variance. TLM-RL and X-RIC$_{300}$ show intermediate margins (eMBB: 15--16.5~Mbps; URLLC: 6--7~ms), confirming that larger buffers improve performance and consistency. \indent\emph{\underline{\textbf{Resource Efficiency Analysis.}}} Fig.~\ref{fig:a3}(c) and (d) compare vCW and vRB utilization. Equal Allocation operates near maximum capacity ($C_w = 1.0$ for vCW, 11.8--12.0 blocks for vRB), providing no efficiency margin and incurring maximum costs. BC-trained policies, however, reduce resource costs: X-RIC$_{500}$ reduces vCW usage to 0.65 (35\% reduction) and vRB usage to 8.5--9.5 blocks (25\% reduction) while satisfying SLA constraints. TLM-RL and X-RIC$_{300}$ show intermediate reductions (vCW: 0.75 and 0.67; vRB: 9.5--10.5 and 9.0--9.8 blocks, respectively), indicating that dual filtering extracts high-quality examples that guide policies toward efficient allocations as the demonstration buffer grows.

\section{Conclusion}\label{sect:Con}
This paper introduced X-RIC, a novel framework integrating agentic AI into O-RAN architectures to address current RL limitations. The proposed framework features several progressive TLM implementation schemes and a hybrid control architecture for sub-second O-RAN decision-making. We introduced BARX, a domain-specific evaluation framework assessing agentic RAN control across latency, compliance, determinism, and policy impact metrics, and developed automated demonstration collection using dual-filtering behavioral cloning to improve sample efficiency while maintaining production-grade reliability. As a case study, we applied our proposed X-RIC framework to a three network slice system, which involves dynamic configuration and poses significant challenges for traditional RL methods to solve within the near-real-time RIC timescale. Validation on SLA-aware network slicing demonstrates that X-RIC significantly outperforms state-of-the-art baselines. 
\balance
\bibliographystyle{ieeetr} 
\bibliography{ref}

\end{document}